\documentclass[runningheads]{llncs}

\usepackage{eccv}

\usepackage{eccvabbrv}
\usepackage{url}            % simple URL typesetting
\usepackage{booktabs}       % professional-quality tables
\usepackage{amsfonts}       % blackboard math symbols
\usepackage{nicefrac}       % compact symbols for 1/2, etc.
\usepackage{microtype}      % microtypography
\usepackage{xcolor}         % colors
\usepackage{multirow}
\usepackage{graphicx}
\usepackage{soul}
\usepackage{algorithmic}
\usepackage{algorithm}
\definecolor{frenchblue}{rgb}{0.0, 0.45, 0.73}

\usepackage{booktabs}

\usepackage[accsupp]{axessibility}  % Improves PDF readability for those with disabilities.

\usepackage[pagebackref,breaklinks,colorlinks,citecolor=eccvblue]{hyperref}
\usepackage{orcidlink}

\begin{document}

\title{Tendency-driven Mutual Exclusivity for Weakly Supervised Incremental Semantic Segmentation} 

\titlerunning{TME for WILSS}

\author{Chongjie Si \and
Xuehui Wang \and
Xiaokang Yang \and Wei Shen}

\authorrunning{Chongjie.~Author et al.}

\institute{MoE Key Lab of Artificial Intelligence, AI Institute, Shanghai Jiao Tong University\\
\email{\{chongjiesi, wangxuehui, xkyang, wei.shen\}@sjtu.edu.cn}}

\maketitle

\begin{abstract}
Weakly Incremental Learning for Semantic Segmentation (WILSS) leverages a pre-trained segmentation model to segment new classes using cost-effective and readily available image-level labels. A prevailing way to solve WILSS is the generation of seed areas for each new class, serving as a form of pixel-level supervision. However, a scenario usually arises where a pixel is concurrently predicted as an old class by the pre-trained segmentation model and a new class by the seed areas. Such a scenario becomes particularly problematic in WILSS, as the lack of pixel-level annotations on new classes makes it intractable to ascertain whether the pixel pertains to the new class or not. To surmount this issue, we propose an innovative, tendency-driven relationship of mutual exclusivity, meticulously tailored to govern the behavior of the seed areas and the predictions generated by the pre-trained segmentation model. This relationship stipulates that predictions for the new and old classes must not conflict whilst prioritizing the preservation of predictions for the old classes, which not only addresses the conflicting prediction issue but also effectively mitigates the inherent challenge of incremental learning - catastrophic forgetting. Furthermore, under the auspices of this tendency-driven mutual exclusivity relationship, we generate pseudo masks for the new classes, allowing for concurrent execution with model parameter updating via the resolution of a bi-level optimization problem.
Extensive experiments substantiate the effectiveness of our framework, resulting in the establishment of new benchmarks and paving the way for further research in this field.
\end{abstract}    
\section{Introduction \label{sec: intro}}

Incremental learning (IL) emerges as a pivotal pathway for counteracting catastrophic forgetting \cite{mccloskey1989catastrophic, french1999catastrophic, cermelli2022incrementalwilson} in deep learning paradigms. Predominantly in the area of image classification, numerous methodologies have been curated to fortify IL \cite{ wu2018memory,shin2017continual, zhu2021prototype, kirkpatrick2017overcoming, ostapenko2019learning, rebuffi2017icarllwfmc}. Recently, semantic segmentation has witnessed a surge in methods tailored for incremental learning tasks (ILSS) \cite{douillard2021plop, cermelli2020modelingmib, klingner2020classcil, michieli2021knowledge, michieli2021continualsdr}. These methods predominantly fall into regularization-based \cite{douillard2021plop, michieli2021continualsdr, cermelli2020modelingmib} or replay-based \cite{yu2023foundation2023cvpr} categories, with the former prioritizing knowledge retention from previous training phases, and the latter employing additional samples to alleviate catastrophic forgetting. Nonetheless, a persistent hurdle is the dependency of ILSS approaches on pixel-level annotations for new classes, which is a resource-intensive and time-consuming requirement.

To address this limitation, very recently, a novel task, weakly incremental learning for semantic segmentation (WILSS), has been proposed by \cite{cermelli2022incrementalwilson}. In the WILSS paradigm, the initial training of a segmentation model takes place on a set of classes with pixel-level annotations. Thereafter, in the following incremental phases, WILSS shifts its focus exclusively to the utilization of image-level labels for new classes, with no access to old data. A prevalent way to tackle WILSS is the generation of seed areas for new classes to serve as pseudo pixel-level supervision. However, the seed areas are inevitably noisy, which might lead to a quandary: A pixel's label is predicted as belonging to an old class by the previous segmentation model but is simultaneously predicted as a new class by the seed area in the incremental phases. With only image-level labels available for the new classes, a formidable conundrum arises in determining whether this pixel's true label corresponds to the new classes or not. Given this scenario, a critical inquiry surfaces: how might we generate high-quality pseudo pixel-level labels for the new classes to address this conflict issue?

\begin{figure*}
    \centering
    \includegraphics[scale=0.25]{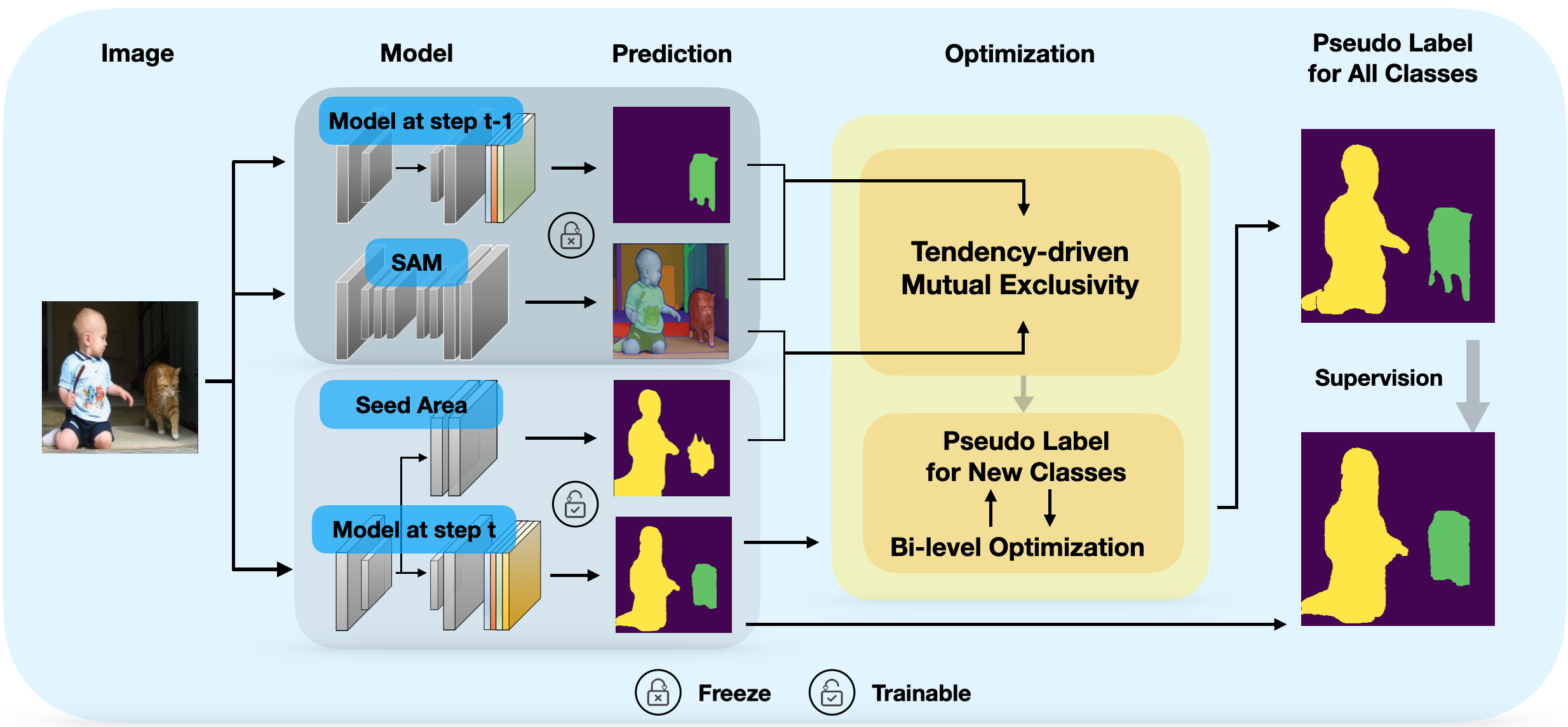}
    \caption{Illustration of the proposed Teddy framework. A novel tendency-driven relationship of mutual exclusivity is proposed to regulate the interactions between the predictions produced by the seed areas and the previous model. Moreover, by solving a TME constrained bi-level optimization problem, we can generate pixel-level pseudo labels for all classes and update model parameters simultaneously.}
    \label{fig:framwork}
\end{figure*}

In this paper, we present Teddy, a novel approach designed to address the aforementioned challenge of conflicting predictions, as shown in Fig. \ref{fig:framwork}. Specifically, the core of Teddy is the establishment of a unique tendency-driven relationship of mutual exclusivity (TME). This relationship is designed to harmonize the predictions between old and new classes. To elaborate, when a pixel is identified as belonging to an old class by the previous segmentation model, it is expected to be treated as background by the seed areas of new classes. TME not only effectively addresses the conflicting prediction issue, but also inherently strengthens the preservation of old class predictions, providing a robust countermeasure to the challenge of catastrophic forgetting.

Within the framework of Teddy, the TME relationship is meticulously crafted as a constraint for the optimization problem intrinsic to WILSS. To establish such constraint, we seek to binarize the predictions for the old classes and then produce predictions for new classes. The nexus between these old and new class predictions is deftly steered by the constraint of TME.
Moreover, we leverage the capabilities of the Segment Anything Model (SAM) \cite{kirillov2023seganySAM}, to further bolsters this process.
Finally, by formulating the optimization objective of WILSS with the TME constraint as a bi-level optimization problem, we are able to produce pixel-level pseudo labels for all classes. This process is executed concurrently with the updating of model parameters, ensuring an efficient and integrated approach.

The contributions of this paper are as follows:

\begin{itemize}
    \item We propose a novel tendency-driven relationship of mutual exclusivity in WILSS, which effectively mitigates the conflict of the predictions generated by the pre-trained model and the seed areas. 
    \item We propose a TME constrained bi-level optimization problem, through which we can simultaneously generate pixel-level pseudo labels for all classes and update the model parameters.
    \item Extensive experiments show that Teddy significantly outperforms state-of-the-art approaches in existing scenarios and data sets, establishing new benchmarks.
        
\end{itemize}

\section{Related Work}

\subsection{Weakly Supervised Semantic Segmentation}
Collecting precise pixel-wise annotations for fully supervised semantic segmentation is a resource-intensive and costly endeavor. In response to this challenge, researchers have introduced weakly-supervised semantic segmentation (WSSS) methods, which aims to train highly effective segmentation models using more cost-efficient forms of supervision, such as bounding boxes \cite{oh2021background, dai2015boxsup, khoreva2017simple, papandreou2015weakly}, scribbles \cite{lin2016scribblesup, vernaza2017learning, tang2018normalized} and image-level labels \cite{du2022weakly, fan2020learning, huang2018weakly, jiang2019integral}. Due to its cost-effectiveness and widespread availability, image-level supervision has garnered the highest level of attention among various forms of weak supervision. The major of WSSS methods under image-level supervision \cite{du2022weakly, fan2020learning, huang2018weakly, jiang2019integral, sun2020mining, ahn2019weakly, ahn2018learning} typically adheres to a three-stage learning process: initial Class Activation Map (CAM) \cite{zhou2016learning} generation, followed by pseudo-mask generation, and the training of the segmentation model. 

In contrast to the rapid advancements in pseudo-label generation techniques based on image-level supervision, the WSSS models learn from a predetermined set of classes, which can only tackle static scenario. Due to the growing attention for more intricate and demanding scenario of incremental learning, in this paper, we aim to extend a pre-trained segmentation model solely using image-level labels while accommodating the introduction of new classes over time.

\subsection{Incremental Learning Semantic Segmentation}
Incremental learning is devised to address the issue of catastrophic forgetting \cite{mccloskey1989catastrophic, french1999catastrophic, cermelli2022incrementalwilson}, which transpires when a model, in the process of acquiring knowledge regarding new classes over time, gradually loses its capacity to recall previously learned classes. Although an extensive investigation of incremental learning (IL) for image classification has been thoroughly conducted \cite{ostapenko2019learning, rebuffi2017icarllwfmc, wu2018memory,shin2017continual, zhu2021prototype, kirkpatrick2017overcoming, zenke2017continual, castro2018end, chaudhry2018riemannian, dhar2019learning}, there exists only a limited number of methodologies proposed for addressing the incremental learning task for semantic segmentation (ILSS) \cite{douillard2021plop, cermelli2020modelingmib, klingner2020classcil, michieli2021knowledge, michieli2021continualsdr}. The prevailing methods in ILSS can be broadly categorized into two groups, namely regularization-based and replay-based \cite{yu2023foundation2023cvpr}. Regularization-based methods \cite{douillard2021plop, michieli2021continualsdr, cermelli2020modelingmib, roy2023rasp} focus on preserving knowledge from prior training phases. For instance, PLOP \cite{douillard2021plop} introduces a multi-scale pooling method that maintains spatial relationships across long and short distances at the feature level. Moreover, SDR \cite{michieli2021continualsdr} advocates enhancing class-conditional features by minimizing the discrepancy among features belonging to the same class. In replay-based methods, auxiliary samples are generated to counteract catastrophic forgetting. For instance, RECALL \cite{maracani2021recall} tackles the forgetting problem by employing web-crawled images with pseudo labels.

Recently, a novel weakly incremental learning for semantic segmentation (WILSS) task was introduced by WILSON \cite{cermelli2022incrementalwilson}, wherein a pretrained segmentation model is employed to segment novel classes only with image-level labels, and without access to old data. As mentioned in Sec. \ref{sec: intro}, compared to ILSS, the problem of conflicting prediction becomes more pronounced in WILSS because of the absence of valid pixel-level supervisions. To address this issue, in this paper we aims to form a tendency-driven relationship of mutual exclusivity between the old and new predictions, which effectively tackles this issue and mitigates the catastrophic forgetting at the same time.

\subsection{Segment Anything Model}
Recently, Segment Anything Model (SAM) \cite{kirillov2023segment}, which is trained with 1B mask annotations and designed to distinguish a foreground region from a background one through some visual prompts (point, box, etc), has been proposed. SAM has surprising capabilities to segment objects accurately even if the objects are not involved in the training data set, which is ideally meaningful to produce class-agnostic masks for segmentation tasks. While SAM has been widely used \cite{wang2023caption, wang2023scaling, liu2023interngpt, chen2023segmentweaksam} in a relatively short period of time, it does not match the expected level of performance in many downstream tasks \cite{yu2023h2rbox, tang2023can, ma2023segment}. Therefore, how to harness the performance of SAM more effectively and imbue it with semantic information is a question worthy of profound consideration.
\section{Method}

Suppose $\mathcal{X}$ represents the input space, where each image $\mathbf{x} \in \mathcal{X}$ consists of a set of pixels denoted as $\mathcal{I}$, with a cardinality of $|\mathcal{I}| = H \times W = N$, without any loss of generality. Denote $\mathcal{Y}$ the label space. 
In the incremental segmentation setting, the training procedure is organized into multiple sequential steps. At each step $t$ ($t>0$), we observe new classes $\mathcal{C}^t$, consequently expanding the label set to $\mathcal{Y}^t = \mathcal{Y}^{t-1} \cup \mathcal{C}^{t}$. Diverging from the original setting, in WILSS, pixel-level annotations are exclusively provided during the initial step ($t=0$). Subsequently, in the ensuing incremental steps, only cost-effective image-level labels are available, and access to prior training images is restricted. 
The primary objective of WILSS revolves around the adaptation of a model $f_{\theta}: \mathcal{X} \rightarrow \mathbb{R}^{|\mathcal{Y}|\times H \times W}$, which maps the input space to the pixel-level probability label space. Here, $\theta$ represents the parameters. The goal is to equip this model with the capacity to perform segmentation on new classes while retaining knowledge of the previously learned classes without incurring catastrophic forgetting.

\subsection{Overview}
In WILSS, the primary objective during step $t$ ($t>0$) revolves around obtaining the optimal model parameters, symbolized as $\theta^t$. To formalize this optimization task, we begin with the following loss function $\mathcal{L}$:
\begin{equation}
   \min_{\theta^t}\quad \mathcal{L}(\theta^t|\theta^{t-1}; (\mathbf{x}^t, \mathbf{y}^t)).
    \label{eq: simple loss}
\end{equation}
Here, $\theta^{t-1}$ represents the parameters of the model at $t-1$ step, $\mathbf{x}^t$ denotes an image observed at step $t$ and $\mathbf{y}^t\in\{0,1\}^{|\mathcal{C}^t|}$ refers to the corresponding image-level label vector of that image. It is imperative to note that in the incremental steps, our access is confined solely to image-level labels and predictions for the previously learned classes produced by $f_{\theta^{t-1}}$. As discussed in the introduction, it is possible for $f_{\theta^{t-1}}$ to predict a pixel's label as belonging to a previously learned class, while the seed area concurrently predicts it as a new class.

To address this conflicting prediction issue, our proposed method, Teddy, aims to establish a novel tendency-driven relationship of mutual exclusivity (TME) between the seed areas representing new classes and the predictions for the old classes. To form the relationship, we first binarize the predictions of the old classes (Sec \ref{sec: old refine}) which will used to form the TME constraint, and then generate seed areas for the new classes (Sec. \ref{sec: new generation}). Subsequently, we elucidate how to form the TME and how it guides seed area refinement for the new classes (Sec. \ref{sec: tme overall}) as well as final pseudo pixel-level label generation for all the classes (Sec. \ref{sec: final pseudo generation}). Finally, we demonstrate how to optimize the objective function with the constraint of TME (Sec. \ref{sec: bi-level}).

\subsection{Tendency-driven Mutual Exclusivity}

\subsubsection{Prediction Binarization for Old Classes \label{sec: old refine}}

Capitalizing on the fact that the model $f_{\theta^{t-1}}$ has been trained with pixel-level annotations, we anticipate that its prediction for a given image $\mathbf{x}^t$, denoted as $f_{\theta^{t-1}}(\mathbf{x}^t) \in \mathbb{R}^{|\mathcal{Y}^{t-1}|\times H\times W}$, is more reliable than seed areas generated from image-level labels. We proceed to convert $f_{\theta^{t-1}}(\mathbf{x}^t)$ to its one-hot presentation $\delta(f_{\theta^{t-1}}(\mathbf{x}^t))\in\{0,1\}^{|\mathcal{Y}^{t-1}|\times H\times W}$, where $\delta(\cdot)$ represents the \textit{one-hot} function. Furthermore, we harness the power of SAM to generate $m$ binary masks, each represented as $\mathbf{M}_i\in \{0,1\}^{H\times W}, i=1,2,...,m$, to obtain the binarized predictions for old classes, which will be served as a part of TME. Specifically, $\mathbf{M}_i$ is considered as belonging to the $c$-th class if and only if
\begin{equation}
   \frac{\sum(\delta(f_{\theta^{t-1}}(\mathbf{x}^t))_c \odot \mathbf{M}_i)}{\min\left(\sum(\delta(f_{\theta^{t-1}}(\mathbf{x}^t))_c), \sum(\mathbf{M}_i) \right)} > \alpha,
\end{equation}
where $\odot$ denotes the element-wise Hadamard product between two matrices, $\sum(\cdot)$ denotes the sum of all elements in a matrix and $\alpha\geq 0.5$ is a hyper-parameter. It suggests that if the intersection area exceeds $\alpha$ times that of either set, mask $\mathbf{M}_i$ is regarded as belonging to the $c$-th class. To guarantee that $\mathbf{M}_i$ is classified into a single class, we set $\alpha\geq 0.5$. This prevents conflicts arising from masks overlapping with two old classes’ predictions, each covering over $\alpha$, leading to dual classification.

Based on the above assumption, we can have matrix $\mathbf{K} \in \{0,1\}^{m \times |\mathcal{Y}^{t-1}|}$, where ${K}_{ic}=1$ means that mask $\mathbf{M}_i$ belongs to the $c$-th class, otherwise ${K}_{ic}=0$.
Finally, we obtain binarized predictions $\mathbf{R}^{t-1} \in \mathbb{R}^{|\mathcal{Y}^{t-1}|\times H \times W}$ for the old classes as $
    \mathbf{R}^{t-1}_c = \phi(\sum_{i=1}^m K_{ic} \mathbf{M}_i)$,
where $\phi(\cdot)$ represents an element-wise operator, defined as $\phi(a) = \min(1,a)$ for a scalar $a$. To simplify notation, we introduce the function $R(\cdot)$ to denote this binarization process, i.e., $\mathbf{R}^{t-1} = R(f_{\theta^{t-1}}(\mathbf{x}^t), \alpha)$.

\subsubsection{Seed Area Generation for New Classes \label{sec: new generation}}

Drawing inspiration from the existing literature on WSSS \cite{du2022weakly, fan2020learning, huang2018weakly, jiang2019integral, sun2020mining, ahn2019weakly, ahn2018learning}, we can generate seed areas to produce pixel-level pseudo-supervision based on image-level labels for the new classes. The seed area generation process utilizes the features of the encoder within the segmentation model. For a given image $\mathbf{x}^t$ along with its corresponding image-level label $\mathbf{y}^t$, the \emph{de facto} seed area generation method is class activation mapping (CAM) \cite{zhou2016learning}, which outputs a score map $S(\mathbf{x}^t, \mathbf{y}^t) \in \mathbb{R}^{|\mathcal{Y}^t|\times H \times W}$ for the image $\mathbf{x}^t$. Here following \cite{cermelli2022incrementalwilson}, we apply a function $\gamma(\cdot)$, which is a normalized Global Weighted Pooling \cite{cermelli2022incrementalwilson} with the focal penalty term \cite{cermelli2022incrementalwilson} which can better identify all visible parts of objects, on the score map to obtain the image-level prediction $\gamma(S(\mathbf{x}^t, \mathbf{y}^t)) \in \mathbb{R}^{|\mathcal{C}^t|}$. We then proceed to train the seed area by minimizing the classification loss $\mathcal{L}_{cls}$:
\begin{equation}
    \min_{\theta^t} \quad \mathcal{L}_{cls} = \frac{1}{|\mathcal{C}^t|} \sum_{c\in \mathcal{C}^t}\mathcal{L}_{{\rm b}} \left\{\mathbf{y}^t_c, \sigma(\gamma(S(\mathbf{x}^t, \mathbf{y}^t)))_c\right\}.
    \label{eq lcls}
\end{equation}
Here, $\sigma(\cdot)$ represents the logistic function and $\mathcal{L}_{{\rm b}}$ corresponds to the element-wise binary cross-entropy loss. Additionally, to inform seed area about the presence of old classes within the image and guide its attention towards alternative regions, we introduce another supervision provided by the segmentation model $f_{\theta^{t-1}}$, and we aim to minimize the following objective function $\mathcal{L}_{loc}$:
\begin{align}
        \min_{\theta^t}\quad \mathcal{L}_{loc} =  & \frac{1}{|\mathcal{Y}^{t-1}|} \sum_{c\in \mathcal{Y}^{t-1}} \mathcal{L}_{{\rm b}}\left\{\sigma(f_{\theta^{t-1}}(\mathbf{x}^t))_c,  \sigma(S(\mathbf{x}^t, \mathbf{y}^t))_c\right\}.
        \label{eq lloc}
\end{align}

\subsubsection{TME Constraint\label{sec: tme overall}}

\begin{figure}
    \centering
    \includegraphics[scale=0.36]{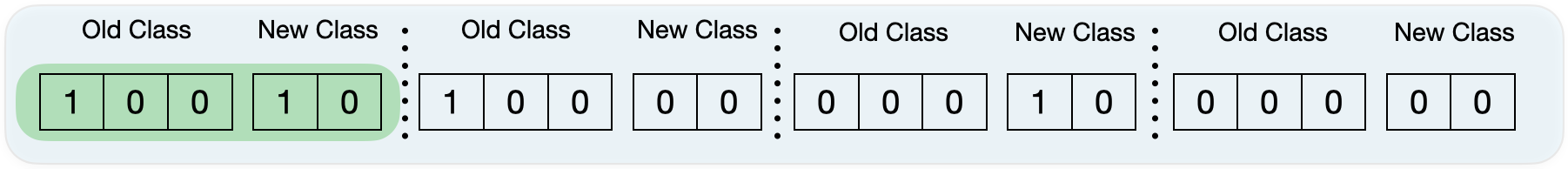}
    \caption{A simple illustration for tendency-driven mutual exclusivity. In each case, the left vector represents $R(f_{\theta^{t-1}}(\mathbf{x}^t),\alpha)_i$, and the right one represents $\delta(S(\mathbf{x}^t, \mathbf{y}^t))_i$. TME will prevent the case under the green shadow while allowing the other three to occur.}
    \label{fig:mutual exclusivity}
\end{figure}
As previously discussed, a pixel may be predicted as foreground by both $f_{\theta^{t-1}}(\mathbf{x}^t)$ and $S(\mathbf{x}^t, \mathbf{y}^t)$ in WILSS. To counteract this, as shown in Fig. \ref{fig:mutual exclusivity}, we aim to establish a relationship of mutual exclusivity between predictions for old and new classes. Specifically, we require that 
\begin{equation}
\begin{aligned}
    &\|R(f_{\theta^{t-1}}(\mathbf{x}^t),\alpha)_i\| + \|\delta(S(\mathbf{x}^t, \mathbf{y}^t))_i\| \leq 1, \\
    &\forall i=(h,w)\in \mathcal{I}.
    \end{aligned}
\end{equation}
This constraint stipulates that for a given pixel, in the predictions of both new and old classes, at most one class is designated as foreground.

\begin{figure}
    \centering
    \includegraphics[scale=0.3]{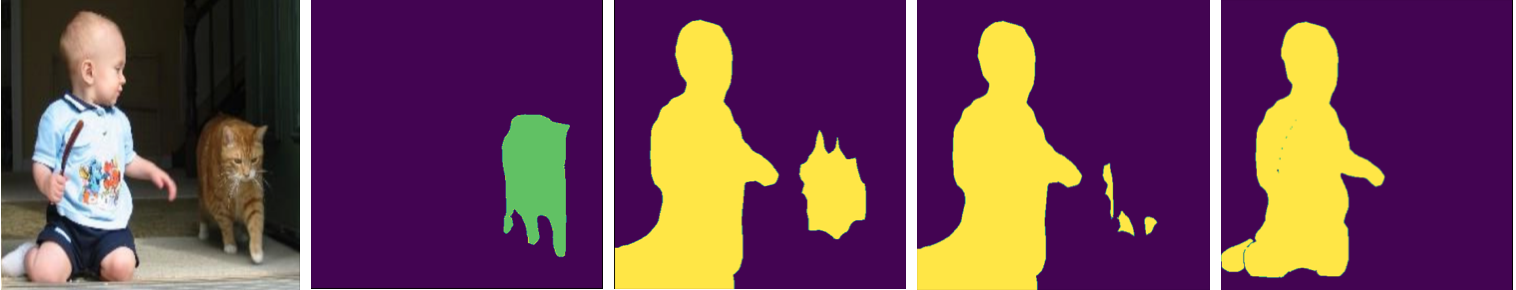}
    \caption{Qualitative results to show the effectiveness of TME. From left to right: image, binarized predictions for the old classes, seed areas without TME for the new classes, seed areas with TME for the new classes and refined seed areas with TME.}
    \label{fig:quality results tme}
\end{figure}

Delving deeper into the implications of this constraint, when a pixel is classified as an old class by $R(f_{\theta^{t-1}}(\mathbf{x}^t),\alpha)$, it is compelled to be classified as background by $\delta(S(\mathbf{x}^t, \mathbf{y}^t))$. In essence, the assignment of new labels is exclusively reserved for pixels predicted as background in the previous step. This constraint significantly preserves the predictions of the old classes and effectively mitigates the issue of conflicting prediction. Furthermore, the establishment of this mutual exclusive relationship hinges upon the predictions of the old classes, rendering it a tendency-driven form of mutual exclusivity. Moreover, Fig. \ref{fig:quality results tme} and the discussion in Sec. \ref{sec: exp tme} demonstrate that the utilization TME is effective in addressing the issue of catastrophic forgetting. This relationship results in the generation of more accurate seed areas for the new classes, showcasing its potential to improve the quality and robustness of the model's performance when learning incrementally. 

\subsection{Pseudo Label Generation for All Classes \label{sec: final pseudo generation}}

We proceed with the generation of pseudo masks for the new classes under the constraint of TME. Following \cite{cermelli2022incrementalwilson}, we can directly derive pseudo labels $\mathbf{P}\in\mathbb{R}^{|\mathcal{C}^t|\times H\times W}$ for the new classes, which is combination of one-hot and softmax forms of the seed areas $S(\mathbf{x}^t, \mathbf{y}^t)$. Subsequently, we can further utilize SAM to produce predictions for the new classes. These predictions are denoted as $R(S(\mathbf{x}^t, \mathbf{y}^t), \beta) \in \mathbb{R}^{|\mathcal{C}^t|\times H\times W}$, where $\beta\geq 0.5$ serves as a hyper-parameter. To fully amalgamating the advantages of both SAM and CAM, we fuse their predictions under the constraint of TME as
\begin{equation}
    \begin{aligned}
    \mathbf{Z} &=\mathbf{U} \odot R(S(\mathbf{x}^t, \mathbf{y}^t), \beta) + \mathbf{V} \odot \mathbf{P}\\
    {\rm s.t. } &\quad \|R(f_{\theta^{t-1}}(\mathbf{x}^t),\alpha)_i\| + \|\delta(S(\mathbf{x}^t, \mathbf{y}^t))_i\| \leq 1, \\
    & \quad \forall i=(h,w)\in \mathcal{I}.
    \end{aligned}
    \label{eq pre z}
\end{equation}
Here, $\mathbf{U},\mathbf{V}\in \mathbb{R}^{|\mathcal{C}^t|\times H\times W}$ are two coefficient matrices that demands optimization, which will be discussed later.

To mitigate the risk of catastrophic forgetting, we use $f_{\theta^{t-1}}(\mathbf{x}^t)$ as the pixel-level pseudo label for the old classes. This substitution is made to counteract any potential bias introduced towards the new classes by the fused pseudo mask. The final pixel-level pseudo-label $\mathbf{G}$ is defined as:
\begin{equation}
    \mathbf{G}_{c}=\left\{\begin{array}{ll}\min \left(\sigma(f_{\theta^{t-1}}(\mathbf{x}^t))_c, \mathbf{Z}_c\right) & \text { if } c=\mathrm{b}, \\
    \mathbf{Z}_c & \text { if } c \in \mathcal{C}^{t}, \\
\sigma\left(f_{\theta^{t-1}}(\mathbf{x}^t)\right)_c &  \text { if } c \in \mathcal{Y}^{t-1}, 
\end{array}\right.
\label{eq final super}
\end{equation}
where $\mathrm{b}$ represents the background class.
%To simplify notation, we introduce the function $H(\cdot)$ to represent the generation of the final pseudo-supervision $\mathbf{G}$ under the constraint of TME as $\mathbf{G} = H\left(\mathbf{x}^t,\mathbf{y}^t; \theta^{t-1} | \theta^t\right)$, and then 
We can then define the segmentation loss $\mathcal{L}_{seg}$ as:
\begin{align}
        \min_{\theta^t, \mathbf{U}, \mathbf{V}}  & \quad\mathcal{L}_{seg}= \frac{1}{|\mathcal{Y}^t|}\sum_{c\in\mathcal{Y}^t} \mathcal{L}_{{\rm b}} \left\{ \mathbf{G}_c, \sigma(f_{\theta^t}(\mathbf{x}^t))_c\right\}\nonumber\\
        {\rm s.t. } & \quad \|R(f_{\theta^{t-1}}(\mathbf{x}^t),\alpha)_i\| + \|\delta(S(\mathbf{x}^t,\mathbf{y}^t))_i\| \leq 1,\nonumber \\
   & \quad \forall i=(h,w)\in \mathcal{I}.
   \label{eq lseg}
\end{align}

\iffalse
\begin{align}
        \min_{\theta^t, \mathbf{U}, \mathbf{V}}  & \frac{1}{|\mathcal{Y}^t|}\sum_{c\in\mathcal{Y}^t} \mathcal{L}_{{\rm b}} \left\{ H\left(\mathbf{x}^t,\mathbf{y}^t; \theta^{t-1} | \theta^t\right)_c, \sigma(f_{\theta^t}(\mathbf{x}^t))_c\right\}\nonumber\\
        {\rm s.t. } & \quad \|R(f_{\theta^{t-1}}(\mathbf{x}^t),\alpha)_i\| + \|\delta(S(\mathbf{x}^t,\mathbf{y}^t))_i\| \leq 1,\nonumber \\
   & \quad \forall i=(h,w)\in \mathcal{I}.
\end{align}
\fi

\subsection{TME Constrained Bi-level Optimization \label{sec: bi-level}}

Based on the preceding discussions, the objective function in Eq. (\ref{eq: simple loss}) is finally formulated as 
\iffalse
\begin{equation}
\begin{aligned}
   \min_{\theta^t, \mathbf{U}, \mathbf{V}} & \mathcal{L}(\theta^t|\theta^{t-1}; (\mathbf{x}^t, \mathbf{y}^t)) \\
    &= \frac{1}{|\mathcal{C}^t|} \sum_{c\in \mathcal{C}^t}\mathcal{L}_{{\rm b}} \left\{\mathbf{y}^t_c, \sigma(\gamma(S(\mathbf{x}^t, \mathbf{y}^t)))_c\right\}\\
   &+\frac{1}{|\mathcal{Y}^{t-1}|} \sum_{c\in \mathcal{Y}^{t-1}} \mathcal{L}_{{\rm b}}\left\{\sigma(f_{\theta^{t-1}}(\mathbf{x}^t))_c,  \sigma(S(\mathbf{x}^t, \mathbf{y}^t))_c\right\}\\
   &+\frac{1}{|\mathcal{Y}^t|}\sum_{c\in\mathcal{Y}^t} \mathcal{L}_{{\rm b}} \left\{ H\left(\mathbf{x}^t,\mathbf{y}^t; \theta^{t-1} |\theta^t\right)_c, \sigma(f_{\theta^t}(\mathbf{x}^t))_c\right\}\\
   {\rm s.t. } & \quad \|R(f_{\theta^{t-1}}(\mathbf{x}^t),\alpha)_i\|_1 + \|\delta(S(\mathbf{x}^t, \mathbf{y}^t))_i\|_1 \leq 1, \\
   & \quad \forall i=(h,w)\in \mathcal{I}.
   \label{eq: final loss}
\end{aligned}
\end{equation}
\fi
\begin{equation}
\begin{aligned}
   \min_{\theta^t, \mathbf{U}, \mathbf{V}} &\quad \mathcal{L}(\theta^t|\theta^{t-1}; (\mathbf{x}^t, \mathbf{y}^t)) = \mathcal{L}_{cls} + \mathcal{L}_{loc} + \mathcal{L}_{seg} \\
   {\rm s.t. } & \quad \|R(f_{\theta^{t-1}}(\mathbf{x}^t),\alpha)_i\| + \|\delta(S(\mathbf{x}^t, \mathbf{y}^t))_i\| \leq 1, \\
   & \quad \forall i=(h,w)\in \mathcal{I}.
   \label{eq: final loss}
\end{aligned}
\end{equation}
During training, we apply the TME constraint by converting the value of $S(\mathbf{x}^t, \mathbf{y}^t)_i$ into a zero vector if $\|R(f_{\theta^{t-1}}(\mathbf{x}^t),\alpha)_i\|=1$ for any $i=(h,w)\in \mathcal{I}$. It is also imperative to determine the values of $\mathbf{U}$ and $\mathbf{V}$ with optimizing the network's parameters. We propose treating $\mathbf{U}$ and $\mathbf{V}$ as latent variables to be optimized concurrently with $\theta^t$. By considering $\mathcal{L}$ as a function of $\theta^t$, $\mathbf{U}$, and $\mathbf{V}$, we formulate the following bi-level optimization problem:
\begin{equation}
    \begin{aligned}
        &\arg\min_{\theta^t}\quad\mathcal{L}(\theta^t, \mathbf{U}^{\ast}, \mathbf{V}^{\ast}); \\
        &\text{subject to }  \|R(f_{\theta^{t-1}}(\mathbf{x}^t),\alpha)_i\| + \|\delta(S(\mathbf{x}^t, \mathbf{y}^t))_i\| \leq 1,\\
        & \qquad \qquad\forall i=(h,w)\in \mathcal{I},\\
        & \qquad \qquad\mathbf{U}^{\ast}, \mathbf{V}^{\ast} = \arg\min_{\mathbf{U}, \mathbf{V}}\mathcal{L}(\theta^t, \mathbf{U}, \mathbf{V}).
    \end{aligned}
    \label{eq: bi-level optimization}
\end{equation}
It is essential to note that bi-level optimization problem has been proven strongly NP-hard \cite{jeroslow1985polynomial}. Consequently, obtaining an exact solution for the problem (\ref{eq: bi-level optimization}) in polynomial time is impossible. Fortunately, the inner optimization problem in Eq. (\ref{eq: bi-level optimization}) has a closed-form solution, with the solution 
\begin{equation}
    \begin{split}
        \mathbf{U}_c &= \phi\left(\mathbb{I}(f_{\theta^t}(\mathbf{x}^t)_c >0 \vee R(S(\mathbf{x}^t, \mathbf{y}^t), \beta)_c \leq \mathbf{P}_c)\right), \\
        \mathbf{V}_c &= \phi\left(\mathbb{I}(f_{\theta^t}(\mathbf{x}^t)_c >0 \vee R(S(\mathbf{x}^t, \mathbf{y}^t), \beta)_c > \mathbf{P}_c)\right).
    \end{split}
    \label{eq: uv solution zheng}
\end{equation}
Here, $\mathbb{I}(\cdot)$ represents the element-wise indicator function. For the detailed explanation on the solution process for $\mathbf{U}$ and $\mathbf{V}$, please refer to Supp. Sec. \ref{sec:sup solution}. 
Therefore, we employ an alternative optimization strategy to approximate the solution, in which $\mathbf{U}$ and $\mathbf{V}$ could be updated in an on-the-fly manner during each optimization iteration.

Moreover, the solution of $\mathbf{U}$ and $\mathbf{V}$ holds practical significance: when encountering a positive model prediction, an amalgamation of both masks occurs. However, if their cumulative value is minuscule, it's inferred as a model misjudgment. Conversely, for a negative model prediction, the lesser of the two masks is designated as the supervisory cue. If the smaller mask's value retains substantial magnitude, there exists room for further honing the model's prediction. This strategy weaves a symbiotic relationship between the model's forecasts and the dual-mask priors, culminating in the amplification of the model's performance.

The complete procedure of Teddy is shown in Alg. \ref{alg} in the Supp.

\section{Experiments}
\subsection{Data sets}
We present an extensive evaluation of Teddy on the Pascal VOC 2012 \cite{everingham2010pascalvoc} and MS COCO \cite{caesar2018coco}. Pascal VOC 2012 contains 1,464 training images and 1,449 validation ones with 20 classes. Following the standard method \cite{cermelli2022incrementalwilson, kolesnikov2016seed}, Pascal VOC data set can be augmented with images from \cite{hariharan2011semanticvocaug}, leading to 10,582 images for training and 1,449 ones for validation. MS COCO is a large-scale data set consisting over 164K images across 80 classes. We also follow the training split and the annotation of \cite{caesar2018coco} to deal with the overlapping annotation problem \cite{lin2014microsoftcoco}. 

\subsection{Protocols}
Following previous works \cite{maracani2021recall, cermelli2020modelingmib, cermelli2022incrementalwilson}, two different incremental learning protocols are introduced: (1) \textbf{disjoint}, where images of each training step only contain current or previous seen classes; (2) \textbf{overlap}, where images of each training step contain any class, among which at least one is new. Obviously, the overlap scenario is more practical and challenging to the disjoint one, as in a real setting there isn’t any oracle method to exclude future classes from the background. Hence, our experiments are primarily concentrated on the overlap setting. However, additional results pertaining to the disjoint setting are available in the supplementary materials.

We start to evaluate Teddy under two widely used \textbf{single-step} incremental learning scenarios: \textbf{15-5 VOC} setting, where 15 classes are learned in the pretraining step and another 5 classes are added in the second step, and \textbf{10-10 VOC} setting, in which 10 classes are first learned with another 10 being learned later. We also consider a more challenging single-step scenario, \textbf{COCO-to-VOC} \cite{cermelli2022incrementalwilson}. Specifically, 60 classes of MS COCO which are not presented in VOC data set is learned in the first step, Afterwards, we learn 20 classes of Pascal VOC. Furthermore, we also consider \textbf{multi-step} settings: \textbf{10-2 VOC} setting, where 2 new classes are learned in 5 incremental steps and \textbf{10-5 VOC} setting, where 5 new classes are learned in each 2 incremental step. Following previous works \cite{maracani2021recall, cermelli2020modelingmib, cermelli2022incrementalwilson}, we report the results on the validation data set since the test set labels have not been publicly released.

\subsection{Baselines}
We compare our method with both supervised incremental learning and weakly supervised semantic segmentation methods, for WILSS is a novel setting proposed by WILSON \cite{cermelli2022incrementalwilson}. For supervised incremental learning methods with pixel-wise annotations, the performance of eight representative state-of-the-art methods including LWF \cite{li2017learninglwf}, LWF-MC \cite{rebuffi2017icarllwfmc}, ILT \cite{michieli2019incrementalilt}, MIB \cite{cermelli2020modelingmib}, PLOP \cite{douillard2021plop}, CIL \cite{klingner2020classcil}, SDR \cite{michieli2021continualsdr}, and RECALL \cite{maracani2021recall}, are compared with.  Moreover, following \cite{cermelli2020modelingmib, maracani2021recall, cermelli2022incrementalwilson}, we also compare with another two methods denoted as FT and Joint. FT is a simple fine-tuning approach, deciding the lower bound of an incremental model, and Joint is trained on all the classed in one step, which can be served as the upper bound. We refer the readers to \cite{cermelli2020modelingmib} for more details on FT and Joint.
As for weakly supervised semantic segmentation methods adopted to the incremental learning scenario, we compare Teddy with the reported results of the pseudo labels generated from four WSSS methods, including CAM \cite{zhou2016learning}, SEAM \cite{wang2020selfseam}, SS \cite{araslanov2020singless} and EPS \cite{lee2021railroadeps}.

\begin{table}[ht]
    \centering
    \caption{Results on different overlap settings. ``P'' indicates pixel-level labels and ``I'' indicates image-level ones. The best method of utilizing image-level supervision is bold, and the best method using pixel level supervision is underlined. FT is a simple fine-tuning approach, deciding the lower bound of an incremental model, and Joint is trained on all the classed in one step, which can be served as the upper bound.}
    \resizebox{\textwidth}{!}{%
    
    \begin{tabular}{l c | c c c | c c c | c c c | c c c | c c c }
    % \hline\hline
    \toprule
         \multirow{3}{*}{Method} & \multirow{3}{*}{Sup} & \multicolumn{9}{c|}{\textbf{Single-step}} & \multicolumn{6}{c}{\textbf{Multi-step}} \\\cline{3-17}
         & & \multicolumn{3}{c|}{\textbf{15-5 VOC}} & \multicolumn{3}{c|}{\textbf{10-10 VOC}} & \multicolumn{3}{c|}{\textbf{COCO-to-VOC}} & \multicolumn{3}{c|}{\textbf{10-2 VOC}} & \multicolumn{3}{c}{\textbf{10-5 VOC}} \\ 
         & & 1-15 & 16-20 & All & 1-10 & 11-20 & All & 1-60 & 61-80 & All & 1-10 & 11-20 & All & 1-10 & 11-20 & All \\ 
         
         \hline
         
         FT & P &  12.5 & 36.9 & 18.3 & 7.8 & 58.9 & 32.1 & 1.9 & 41.7 & 12.7 & - & - & - & - & - & -\\ \hline
         
         LWF \cite{li2017learninglwf} & P & 67.0 & 41.8 & 61.0  & \underline{70.7} & 63.4 & 67.2 & 36.7 & \underline{49.0} & \underline{40.3} & - & - & - & - & - & -\\
         
         LWF-MC \cite{rebuffi2017icarllwfmc} & P & 59.8 & 22.6 & 51.0 & 53.9 & 43.0 & 48.7  & - & - & - & - & - & - & - & - & - \\

         ILT \cite{michieli2019incrementalilt} & P & 69.0 & 46.4 & 63.6 & 70.3 & 61.9 & 66.3 & \underline{37.0} & 43.9 & 39.3 & - & - & - & - & - & - \\

         CIL \cite{klingner2020classcil} & P & 14.9 & 37.3 & 20.2 & 38.4 & 60.0 & 48.7 & - & - & - & - & - & - & - & - & - \\
 
         MiB \cite{cermelli2020modelingmib} & P & 75.5 & 49.4 & 69.0 & 70.4 & 63.7 & 67.2 & 34.9 & 47.8 & 38.7 & - & - & - & - & - & - \\

         PLOP \cite{douillard2021plop} & P & \underline{75.7} & 51.7 & \underline{70.1} & 69.6 & 62.2 & 67.1 & 35.1 & 39.4 & 36.8 & - & - & - & - & - & - \\

         SDR \cite{michieli2021continualsdr} & P & 75.4 & 52.6 & 69.9 & 70.5 & \underline{63.9} & \underline{67.4} & - & - & - & - & - & - & - & - & - \\

         RECALL \cite{maracani2021recall} & P & 67.7 & \underline{54.3} & 65.6 & 66.0 & 58.8 & 63.7 & - & - & - & - & - & - & - & - & - \\

        \hline
        
         CAM \cite{zhou2016learning} & I & 69.9 & 25.6 & 59.7 & 70.8 & 44.2 & 58.5 & 30.7 & 20.3 & 28.1 & - & - & - & - & - & - \\

         SEAM \cite{wang2020selfseam} & I & 68.3 & 31.8 & 60.4 & 67.5 & 55.4 & 62.7 & 31.2 & 28.2 & 30.5 & - & - & - & - & - & -\\

         SS \cite{araslanov2020singless} & I & 72.2 & 27.5 & 62.1 & 69.6 & 32.8 & 52.5 & 35.1 & 36.9 & 35.5 & - & - & - & - & - & -  \\

         EPS \cite{lee2021railroadeps} & I & 69.4 & 34.5 & 62.1 & 69.0 & 57.0 & 64.3 & 34.9 & 38.4 & 35.8 & - & - & - & - & - & -\\
         
         WILSON \cite{cermelli2022incrementalwilson} & I & 74.2 & 41.7 & 67.2 & 70.4 & 57.1 & 65.0 & 39.8 & 41.0 & 40.6 & 38.7 & 22.4 & 32.5 & 66.8 & 46.5 & 58.1\\
         
         \textbf{Teddy} & I  & \textbf{77.6} & \textbf{51.4} & \textbf{72.0} & \textbf{71.2} & \textbf{59.4} & \textbf{66.5} & 
         \textbf{40.6} & \textbf{41.8} & \textbf{41.5} & \textbf{50.3} & \textbf{32.0} & \textbf{43.1} & \textbf{68.9} & \textbf{51.7} & \textbf{61.7}
         \\\hline
        
        Joint & P & 75.5 & 73.5 & 75.4 & 76.6 & 74.0 & 75.4 & - & - & - & - & - & - & - & - & -\\
         \bottomrule
    \end{tabular}}

    \label{tab: different setting results}
\end{table}

\begin{table}[ht]\footnotesize
    \centering
    \setlength{\tabcolsep}{1.3mm}
    \caption{Ablation study. ``OB'' stands for prediction binarization for old classes, ``TME'' stands for tendency-driven mutual exclusivity and ``PF'' stands for prediction fusion for new classes.}
    \begin{tabular}{c c c c|c c c | c c c | c c c}
    \toprule
    \multirow{2}{*}{Row} & \multicolumn{2}{c}{TME} & \multirow{2}{*}{PF} & \multicolumn{3}{c|}{\textbf{15-5 VOC}} & \multicolumn{3}{c|}{\textbf{10-10 VOC}} & \multicolumn{3}{c}{\textbf{10-2 VOC}} \\
    & OB & w/o OB & & 1-15 & 16-20 & All & 1-10 & 11-20 & All & 1-10 & 11-20 & All\\\hline
    
    1 & & & & 74.1 & 41.5 & 67.0 & 70.4 & 57.0 & 65.0 & 38.8 & 22.3 & 32.5\\
    
    2 &  & & \checkmark & 75.2 & 45.6 & 68.9 & 70.7 & 57.5 & 65.3 & 47.7 & 20.1 & 34.6 \\

    3 & & \checkmark & & 76.8 & 46.0 & 70.1 & 71.0 & 58.6 & 66.0 & 50.2 & 23.7 & 38.5\\

    4 & \checkmark & & & 77.0 & 47.7 & 70.7 & 71.2 & 58.6 & 66.2 & 50.0 & 27.5 & 40.8  \\

    5 & & \checkmark & \checkmark & 76.5 & 48.1 & 70.4 & 71.1 & 58.5 & 66.1 & 51.4 &25.2 & 39.9  \\

    \hline
    
    6 & \checkmark &  & \checkmark & 77.6 & 51.4 & 72.0 & 71.2 & 59.4 & 66.5 & 50.3 & 32.0 & 43.1\\
    
    \bottomrule
    \end{tabular}
    \label{tab:ablation study}
\end{table}

\subsection{Implementation Details}
We employ Deeplab V3 \cite{chen2017rethinkingdeeplabv3} for all the settings, with the backbone of ResNet-101 \cite{he2016deepresnet} for Pascal VOC and Wide-ResNet-38 \cite{wu2019wider} for COCO, both pretrained on ImageNet \cite{deng2009imagenet}. Similar to \cite{cermelli2020modelingmib, cermelli2022incrementalwilson}, we employ in-place activated batch normalization \cite{bulo2018place} to optimize memory usage for our experiments. The model is trained for 40 epochs with the batch size of 24 on two GPUs under distributed data parallel. The optimizer is SGD with a learning rate of 0.001, momentum 0.9 and weight decay $10^{-4}$. Furthermore, the model is first warmed up for five epochs. After that, the model is trained with the learning rate decaying by using a polynomial schedule with a power of 0.9. We set $\alpha=0.8$, $\beta=0.5$ for all the settings. For SAM, we use ViT-H \cite{dosovitskiy2020image} as the backbone by default.

\iffalse
\begin{table}[ht]
    \centering
    \caption{Results on multi-step settings. ``I'' indicates image-level supervision. The best method is bold.}
    
    \begin{tabular}{l c | c c c | c c c | c c c | c c c }
    % \hline\hline
    \toprule
         \multirow{3}{*}{Method} & \multirow{3}{*}{Sup} & \multicolumn{6}{c|}{\textbf{10-2 VOC}} & \multicolumn{6}{c}{\textbf{10-5 VOC}}\\ \cline{3-14}
         & & \multicolumn{3}{c}{Disjoint} & \multicolumn{3}{c|}{Overlap} & \multicolumn{3}{c}{Disjoint} & \multicolumn{3}{c}{Overlap} \\ 
         & & 1-10 & 11-20 & All & 1-10 & 11-20 & All & 1-10 & 11-20 & All & 1-10 & 11-20 & All \\ 
        \hline
         WILSON \cite{cermelli2022incrementalwilson} & I & 36.4 & 20.8 & 30.6 & 38.7 & 22.4 & 32.5 & 58.6 & 45.3 & 53.6 & 66.8 & 46.5 & 58.1\\
         \textbf{Teddy} & I & & & & \textbf{50.3} & \textbf{32.0} & \textbf{43.1} & & & & \textbf{68.9} & \textbf{51.7} & \textbf{61.7}
         \\ \bottomrule
         
    \end{tabular}

    \label{tab: multi setting results}
\end{table}
\fi

\subsection{Results}

\subsubsection{15-5 VOC}
In this experimental setting, we introduce an additional 5 classes from the VOC data set after the initial stage. To ensure a fair comparison, we select the same classes as those in \cite{cermelli2022incrementalwilson}, which include: \textit{plant, sheep, sofa, train, tv monitor}. As demonstrated in Table \ref{tab: different setting results}, Teddy attains a new state-of-the-art performance when compared to baselines trained with either pixel-level or image-level labels. It is noteworthy that Teddy even surpasses fully-supervised methods by 2.7\%. In comparison to weak-supervised methods, Teddy outperforms WILSON by \textbf{7.1\%} in general. Notably, for new classes Teddy outperforms WILSON by \textbf{23.3\%}. Furthermore, Teddy demonstrates enhanced performance for old classes by \textbf{4.6\%}, signifying that the phenomenon of catastrophic forgetting has been effectively mitigated.

\subsubsection{10-10 VOC}
In this setting, 10 classes are introduced in the incremental step: \textit{dining table, dog, horse, motorbike, person, plant, sheep, sofa, train, tv monitor}. As shown in Table \ref{tab: different setting results}, Teddy achieves new SOTA performances and even surpasses some fully-supervised methods. Specifically, Teddy outperforms WILSON by 2.3\% in general, and for new classes Teddy outperforms WILSON by \textbf{4.0\%}.

\subsubsection{COCO-to-VOC}
This setting presents a greater challenge. Initially, the base model is trained on 60 classes from the COCO data set, which do not overlap with the classes in VOC. Subsequently, in the second step, we train the additional 20 classes from the VOC data set using only image-level labels. The results are shown in Table \ref{tab: different setting results}, where we can observe that despite the heightened complexity and challenges intrinsic to this setup, Teddy achieves new SOTA performances. Notably, Teddy effectively preserves the previously acquired knowledge, epitomizing its prowess in counteracting catastrophic forgetting.

\subsubsection{10-2 VOC and 10-5 VOC} In these two settings, the issue of catastrophic forgetting exacerbates in these settings due to the increased number of tasks. Table \ref{tab: different setting results} demonstrates that Teddy markedly surpasses WILSON. Notably, in the 10-2 VOC setting, Teddy exceeds WILSON's performance by a substantial margin of \textbf{32.6\%}, and for the 10-5 VOC setting, Teddy outperforms WILSON by \textbf{6.2\%}.  The TME module effectively helps Teddy preserve the learnt knowledge, significantly reducing the impact of catastrophic forgetting, which leads to a substantial improvement.

\subsection{Ablation Study}

\subsubsection{Tendency-driven Mutual Exclusivity \label{sec: exp tme}}
To assess the contribution of the TME in Teddy, we conduct comparisons between Teddy and that without TME. In this TME-excluded approach, the predictions for the new classes are directly refined, skipping the influence of TME. The comparative outcomes are presented in row 2 in Table \ref{tab:ablation study}. Evidently, comparing with rows 2 and 6, Teddy outperforms its degenerated version by \textbf{3.1, 1.2 and 8.5 mIoU}, respectively. Moreover, when comparing rows 1 and 3, the TME constraint can also boost a common WILSS model by \textbf{3.1, 1.0 and 6.0 mIoU}, respectively.

\begin{table*}[ht]
    \centering
    \setlength{\tabcolsep}{1.3mm}
    \caption{Effectiveness of prediction fusion for new classes through optimization.}
    
    \begin{tabular}{c c|c c c | c c c |c c c}
    \toprule
    \multirow{2}{*}{$\mathbf{U}$} & \multirow{2}{*}{$\mathbf{V}$} & \multicolumn{3}{c|}{\textbf{15-5 VOC}} & \multicolumn{3}{c|}{\textbf{10-10 VOC}} & \multicolumn{3}{c}{\textbf{10-2 VOC}} \\
     & & 1-15 & 16-20 & All & 1-10 & 11-20 & All & 1-10 & 11-20 & All\\\hline
     
     0.25 & 0.75 & 76.8 & 46.7 & 70.3 & 70.6 & 57.4 & 65.2 & 48.8 & 29.0 & 40.9\\
     
     0.50 & 0.50 & 77.0 & 48.1 & 71.1 & 71.0 & 58.2 & 65.8 & 49.1 & 28.8 & 40.8 \\
     
     0.75 & 0.25 & 76.8 & 48.1 & 71.0 & 71.0 & 58.7 & 66.1 & 47.2 & 23.5 & 37.3\\\hline
    \multicolumn{2}{c|}{Optimization}& 77.6 & 51.4 & 72.0 & 71.2 & 59.4 & 66.5 & 50.3 & 32.0 & 43.1\\
    \bottomrule
    \end{tabular}
    \label{tab: mask fusion}
\end{table*}

Augmenting our empirical analysis, we proffer a series of qualitative insights to underscore the potency of TME. As shown in Fig. \ref{fig:quality results tme}, it is obvious that the predictions with TME for the new classes evince greater precision than those without TME. The TME relationship not only encapsulates rich information of old classes but also effectively addresses the conflicting prediction issue and catastrophic forgetting, thereby resulting in superior performance.

Furthermore, the results of Teddy, when executed without prediction binarization for old classes (OB) in TME, are shown in row 5 of Table \ref{tab:ablation study}. Based on the results, it becomes apparent that incorporating prediction binarization into TME elevating the model's performance. By harnessing the prowess of SAM to execute binarization on predictions for old classes, we ensure that this information is effectively and accurately integrated into the TME constraint, which subsequently culminates in markedly improved outcomes.

\subsubsection{Incorporation of SAM}
In Teddy, SAM takes part in the process of old class prediction binarization and new class prediction fusion. We further appraise the usefulness of SAM by proceeding without its incorporation. Comparing rows 3 and 4 in Table \ref{tab:ablation study}, we can find that the old class prediction binarization process can improve Teddy's performance by 0.6, 0.2 and 2.3 mIoU respectively. When comparing row 3 and 5, it is observed that the new class prediction fusion further boosts the performance by 0.3, 0.1 and 1.4 mIoU. This further demonstrates the effectiveness of SAM in these processes.

\subsubsection{Prediction Fusion for New Classes}
Firstly, we evaluate the usefulness of prediction fusion (PF) for new classes. By omitting $R(S(\mathbf{x}^t, \mathbf{y}^t), \beta)$ and directly considering $\mathbf{P}$ under the TME constraint as the final mask for new classes, the performances are shown in row 4 in Table \ref{tab:ablation study}. An evident decrease across different settings underscores the efficacy of prediction fusion.

Furthermore, by configuring $\mathbf{U}$ and $\mathbf{V}$ as constants, we contrast the outcomes generated from various prediction fusion strategies. The results in Table \ref{tab: mask fusion} conspicuously indicate that coefficients derived from the optimization problem yield superior performance compared to other strategies. This observation underscores the value of our prediction fusion strategy in efficaciously merging predictions, thereby enhancing overall performance.

\subsubsection{Qualitative Results}
\begin{figure}[ht]
    \centering
    \includegraphics[scale=0.35]{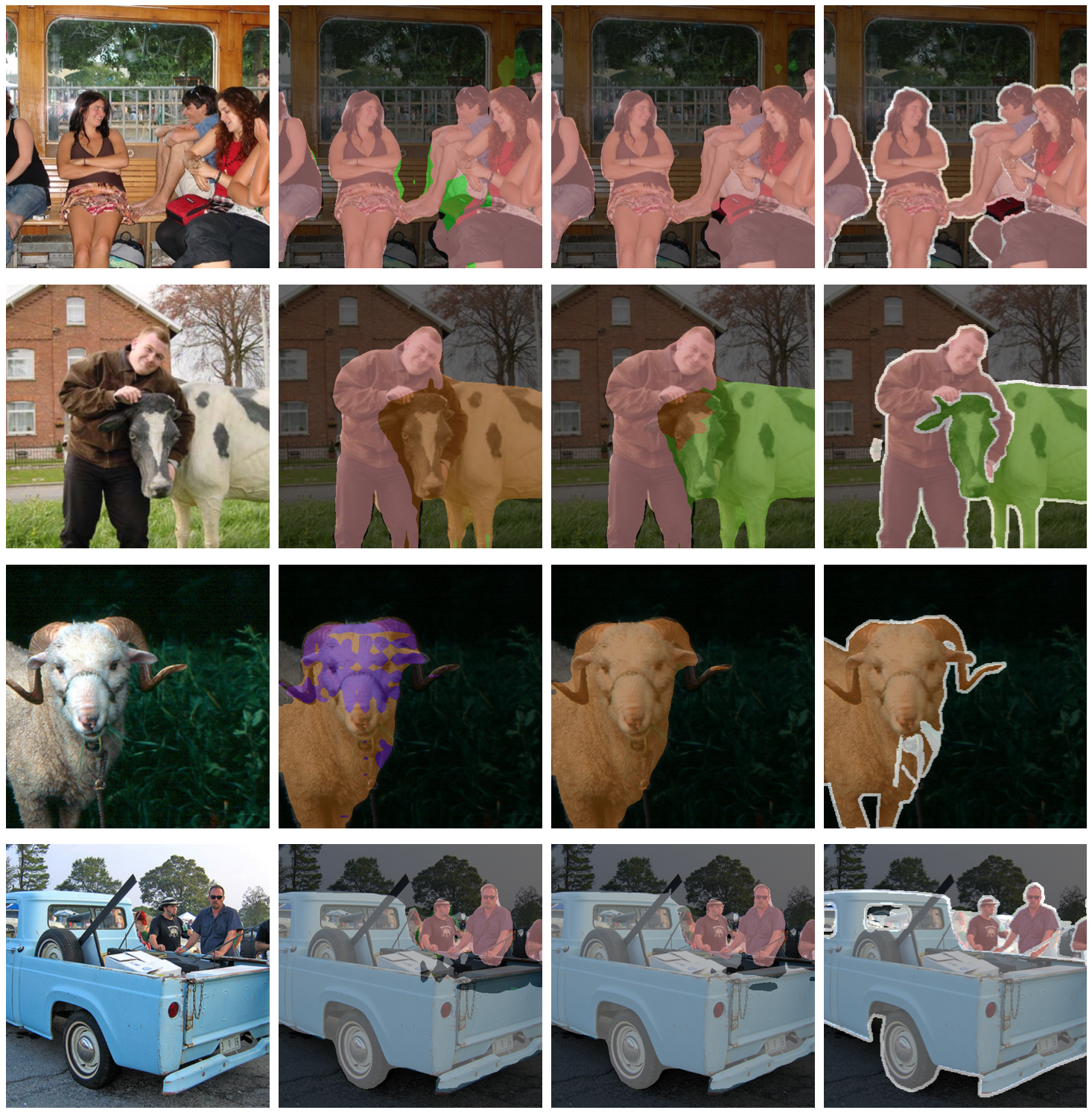}
    \caption{Qualitative results on 15-5 VOC setting for both new (sheep) and old classes. From left to right: image, WILSON \cite{cermelli2022incrementalwilson}, Teddy and the ground-truth.}
    \label{fig:quality}
\end{figure}

In addition, we present qualitative results to further illustrate Teddy's superiority. As depicted in Fig. \ref{fig:quality}, it becomes evident that Teddy outperforms WILSON by producing more precise and well-defined segmentation results. Furthermore, Teddy exhibits accurate predictions for the old classes, serving as validation of its effectiveness in mitigating catastrophic forgetting.

%\subsection{Limitations}
%It's worth noting that Teddy has certain limitations, particularly when it comes to single-class incremental learning. Specifically, Teddy requires images with at least two classes present in order to generate seed areas.
%In the future, it is also interesting to investigate how to perform single-class incremental steps in WILSS.
\section{Conclusion}

In this paper, we introduce a novel framework called Teddy to address the intricate challenges posed by WILSS. Specifically, we propose a novel tendency-driven relationship of mutual exclusivity (TME) to regulate the interactions between the seed area and the predictions generated by the previous model, which not only resolves the conflicting prediction issue but also robustly combats catastrophic forgetting. Harnessing the strength of this relationship, we generate pseudo masks for new classes, allowing for concurrent execution with model parameter updating through a bi-level optimization strategy. Our extensive experiments, conducted on the Pascal VOC and MS COCO data sets, provide convincing evidence of the effectiveness and superiority of our framework, and validate that Teddy successfully addresses the issues of catastrophic forgetting and the conflicting prediction issue in WILSS.

\bibliographystyle{splncs04}
\bibliography{main}

\begin{thebibliography}{10}
\providecommand{\url}[1]{\texttt{#1}}
\providecommand{\urlprefix}{URL }
\providecommand{\doi}[1]{https://doi.org/#1}

\bibitem{ahn2019weakly}
Ahn, J., Cho, S., Kwak, S.: Weakly supervised learning of instance segmentation with inter-pixel relations. In: Proceedings of the IEEE/CVF conference on computer vision and pattern recognition. pp. 2209--2218 (2019)

\bibitem{ahn2018learning}
Ahn, J., Kwak, S.: Learning pixel-level semantic affinity with image-level supervision for weakly supervised semantic segmentation. In: Proceedings of the IEEE conference on computer vision and pattern recognition. pp. 4981--4990 (2018)

\bibitem{araslanov2020singless}
Araslanov, N., Roth, S.: Single-stage semantic segmentation from image labels. In: Proceedings of the IEEE/CVF Conference on Computer Vision and Pattern Recognition. pp. 4253--4262 (2020)

\bibitem{bulo2018place}
Bulo, S.R., Porzi, L., Kontschieder, P.: In-place activated batchnorm for memory-optimized training of dnns. In: Proceedings of the IEEE Conference on Computer Vision and Pattern Recognition. pp. 5639--5647 (2018)

\bibitem{caesar2018coco}
Caesar, H., Uijlings, J., Ferrari, V.: Coco-stuff: Thing and stuff classes in context. In: Proceedings of the IEEE conference on computer vision and pattern recognition. pp. 1209--1218 (2018)

\bibitem{castro2018end}
Castro, F.M., Mar{\'\i}n-Jim{\'e}nez, M.J., Guil, N., Schmid, C., Alahari, K.: End-to-end incremental learning. In: Proceedings of the European conference on computer vision (ECCV). pp. 233--248 (2018)

\bibitem{cermelli2022incrementalwilson}
Cermelli, F., Fontanel, D., Tavera, A., Ciccone, M., Caputo, B.: Incremental learning in semantic segmentation from image labels. In: Proceedings of the IEEE/CVF Conference on Computer Vision and Pattern Recognition. pp. 4371--4381 (2022)

\bibitem{cermelli2020modelingmib}
Cermelli, F., Mancini, M., Bulo, S.R., Ricci, E., Caputo, B.: Modeling the background for incremental learning in semantic segmentation. In: Proceedings of the IEEE/CVF Conference on Computer Vision and Pattern Recognition. pp. 9233--9242 (2020)

\bibitem{chaudhry2018riemannian}
Chaudhry, A., Dokania, P.K., Ajanthan, T., Torr, P.H.: Riemannian walk for incremental learning: Understanding forgetting and intransigence. In: Proceedings of the European conference on computer vision (ECCV). pp. 532--547 (2018)

\bibitem{chen2017rethinkingdeeplabv3}
Chen, L.C., Papandreou, G., Schroff, F., Adam, H.: Rethinking atrous convolution for semantic image segmentation. arXiv preprint arXiv:1706.05587  (2017)

\bibitem{chen2023segmentweaksam}
Chen, T., Mai, Z., Li, R., Chao, W.l.: Segment anything model (sam) enhanced pseudo labels for weakly supervised semantic segmentation. arXiv preprint arXiv:2305.05803  (2023)

\bibitem{dai2015boxsup}
Dai, J., He, K., Sun, J.: Boxsup: Exploiting bounding boxes to supervise convolutional networks for semantic segmentation. In: Proceedings of the IEEE international conference on computer vision. pp. 1635--1643 (2015)

\bibitem{deng2009imagenet}
Deng, J., Dong, W., Socher, R., Li, L.J., Li, K., Fei-Fei, L.: Imagenet: A large-scale hierarchical image database. In: 2009 IEEE conference on computer vision and pattern recognition. pp. 248--255. Ieee (2009)

\bibitem{dhar2019learning}
Dhar, P., Singh, R.V., Peng, K.C., Wu, Z., Chellappa, R.: Learning without memorizing. In: Proceedings of the IEEE/CVF conference on computer vision and pattern recognition. pp. 5138--5146 (2019)

\bibitem{dosovitskiy2020image}
Dosovitskiy, A., Beyer, L., Kolesnikov, A., Weissenborn, D., Zhai, X., Unterthiner, T., Dehghani, M., Minderer, M., Heigold, G., Gelly, S., et~al.: An image is worth 16x16 words: Transformers for image recognition at scale. arXiv preprint arXiv:2010.11929  (2020)

\bibitem{douillard2021plop}
Douillard, A., Chen, Y., Dapogny, A., Cord, M.: Plop: Learning without forgetting for continual semantic segmentation. In: Proceedings of the IEEE/CVF Conference on Computer Vision and Pattern Recognition. pp. 4040--4050 (2021)

\bibitem{du2022weakly}
Du, Y., Fu, Z., Liu, Q., Wang, Y.: Weakly supervised semantic segmentation by pixel-to-prototype contrast. In: Proceedings of the IEEE/CVF Conference on Computer Vision and Pattern Recognition. pp. 4320--4329 (2022)

\bibitem{everingham2010pascalvoc}
Everingham, M., Van~Gool, L., Williams, C.K., Winn, J., Zisserman, A.: The pascal visual object classes (voc) challenge. International journal of computer vision  \textbf{88},  303--338 (2010)

\bibitem{fan2020learning}
Fan, J., Zhang, Z., Song, C., Tan, T.: Learning integral objects with intra-class discriminator for weakly-supervised semantic segmentation. In: Proceedings of the IEEE/CVF Conference on Computer Vision and Pattern Recognition. pp. 4283--4292 (2020)

\bibitem{french1999catastrophic}
French, R.M.: Catastrophic forgetting in connectionist networks. Trends in cognitive sciences  \textbf{3}(4),  128--135 (1999)

\bibitem{hariharan2011semanticvocaug}
Hariharan, B., Arbel{\'a}ez, P., Bourdev, L., Maji, S., Malik, J.: Semantic contours from inverse detectors. In: 2011 international conference on computer vision. pp. 991--998. IEEE (2011)

\bibitem{he2016deepresnet}
He, K., Zhang, X., Ren, S., Sun, J.: Deep residual learning for image recognition. In: Proceedings of the IEEE conference on computer vision and pattern recognition. pp. 770--778 (2016)

\bibitem{huang2018weakly}
Huang, Z., Wang, X., Wang, J., Liu, W., Wang, J.: Weakly-supervised semantic segmentation network with deep seeded region growing. In: Proceedings of the IEEE conference on computer vision and pattern recognition. pp. 7014--7023 (2018)

\bibitem{jeroslow1985polynomial}
Jeroslow, R.G.: The polynomial hierarchy and a simple model for competitive analysis. Mathematical programming  \textbf{32}(2),  146--164 (1985)

\bibitem{jiang2019integral}
Jiang, P.T., Hou, Q., Cao, Y., Cheng, M.M., Wei, Y., Xiong, H.K.: Integral object mining via online attention accumulation. In: Proceedings of the IEEE/CVF international conference on computer vision. pp. 2070--2079 (2019)

\bibitem{khoreva2017simple}
Khoreva, A., Benenson, R., Hosang, J., Hein, M., Schiele, B.: Simple does it: Weakly supervised instance and semantic segmentation. In: Proceedings of the IEEE conference on computer vision and pattern recognition. pp. 876--885 (2017)

\bibitem{kirillov2023seganySAM}
Kirillov, A., Mintun, E., Ravi, N., Mao, H., Rolland, C., Gustafson, L., Xiao, T., Whitehead, S., Berg, A.C., Lo, W.Y., Doll{\'a}r, P., Girshick, R.: Segment anything. arXiv:2304.02643  (2023)

\bibitem{kirillov2023segment}
Kirillov, A., Mintun, E., Ravi, N., Mao, H., Rolland, C., Gustafson, L., Xiao, T., Whitehead, S., Berg, A.C., Lo, W.Y., et~al.: Segment anything. arXiv preprint arXiv:2304.02643  (2023)

\bibitem{kirkpatrick2017overcoming}
Kirkpatrick, J., Pascanu, R., Rabinowitz, N., Veness, J., Desjardins, G., Rusu, A.A., Milan, K., Quan, J., Ramalho, T., Grabska-Barwinska, A., et~al.: Overcoming catastrophic forgetting in neural networks. Proceedings of the national academy of sciences  \textbf{114}(13),  3521--3526 (2017)

\bibitem{klingner2020classcil}
Klingner, M., B{\"a}r, A., Donn, P., Fingscheidt, T.: Class-incremental learning for semantic segmentation re-using neither old data nor old labels. In: 2020 IEEE 23rd international conference on intelligent transportation systems (ITSC). pp.~1--8. IEEE (2020)

\bibitem{kolesnikov2016seed}
Kolesnikov, A., Lampert, C.H.: Seed, expand and constrain: Three principles for weakly-supervised image segmentation. In: Computer Vision--ECCV 2016: 14th European Conference, Amsterdam, The Netherlands, October 11--14, 2016, Proceedings, Part IV 14. pp. 695--711. Springer (2016)

\bibitem{lee2021railroadeps}
Lee, S., Lee, M., Lee, J., Shim, H.: Railroad is not a train: Saliency as pseudo-pixel supervision for weakly supervised semantic segmentation. In: Proceedings of the IEEE/CVF conference on computer vision and pattern recognition. pp. 5495--5505 (2021)

\bibitem{li2023semanticsam}
Li, F., Zhang, H., Sun, P., Zou, X., Liu, S., Yang, J., Li, C., Zhang, L., Gao, J.: Semantic-sam: Segment and recognize anything at any granularity. arXiv preprint arXiv:2307.04767  (2023)

\bibitem{li2017learninglwf}
Li, Z., Hoiem, D.: Learning without forgetting. IEEE transactions on pattern analysis and machine intelligence  \textbf{40}(12),  2935--2947 (2017)

\bibitem{lin2016scribblesup}
Lin, D., Dai, J., Jia, J., He, K., Sun, J.: Scribblesup: Scribble-supervised convolutional networks for semantic segmentation. In: Proceedings of the IEEE conference on computer vision and pattern recognition. pp. 3159--3167 (2016)

\bibitem{lin2014microsoftcoco}
Lin, T.Y., Maire, M., Belongie, S., Hays, J., Perona, P., Ramanan, D., Doll{\'a}r, P., Zitnick, C.L.: Microsoft coco: Common objects in context. In: Computer Vision--ECCV 2014: 13th European Conference, Zurich, Switzerland, September 6-12, 2014, Proceedings, Part V 13. pp. 740--755. Springer (2014)

\bibitem{liu2023interngpt}
Liu, Z., He, Y., Wang, W., Wang, W., Wang, Y., Chen, S., Zhang, Q., Yang, Y., Li, Q., Yu, J., Li, K., Chen, Z., Yang, X., Zhu, X., Wang, Y., Wang, L., Luo, P., Dai, J., Qiao, Y.: Interngpt: Solving vision-centric tasks by interacting with chatgpt beyond language. arXiv preprint arXiv:2305.05662  (2023)

\bibitem{ma2023segment}
Ma, J., Wang, B.: Segment anything in medical images. arXiv preprint arXiv:2304.12306  (2023)

\bibitem{maracani2021recall}
Maracani, A., Michieli, U., Toldo, M., Zanuttigh, P.: Recall: Replay-based continual learning in semantic segmentation. In: Proceedings of the IEEE/CVF International Conference on Computer Vision. pp. 7026--7035 (2021)

\bibitem{mccloskey1989catastrophic}
McCloskey, M., Cohen, N.J.: Catastrophic interference in connectionist networks: The sequential learning problem. In: Psychology of learning and motivation, vol.~24, pp. 109--165. Elsevier (1989)

\bibitem{michieli2019incrementalilt}
Michieli, U., Zanuttigh, P.: Incremental learning techniques for semantic segmentation. In: Proceedings of the IEEE/CVF international conference on computer vision workshops. pp.~0--0 (2019)

\bibitem{michieli2021continualsdr}
Michieli, U., Zanuttigh, P.: Continual semantic segmentation via repulsion-attraction of sparse and disentangled latent representations. In: Proceedings of the IEEE/CVF conference on computer vision and pattern recognition. pp. 1114--1124 (2021)

\bibitem{michieli2021knowledge}
Michieli, U., Zanuttigh, P.: Knowledge distillation for incremental learning in semantic segmentation. Computer Vision and Image Understanding  \textbf{205},  103167 (2021)

\bibitem{oh2021background}
Oh, Y., Kim, B., Ham, B.: Background-aware pooling and noise-aware loss for weakly-supervised semantic segmentation. In: Proceedings of the IEEE/CVF conference on computer vision and pattern recognition. pp. 6913--6922 (2021)

\bibitem{ostapenko2019learning}
Ostapenko, O., Puscas, M., Klein, T., Jahnichen, P., Nabi, M.: Learning to remember: A synaptic plasticity driven framework for continual learning. In: Proceedings of the IEEE/CVF conference on computer vision and pattern recognition. pp. 11321--11329 (2019)

\bibitem{papandreou2015weakly}
Papandreou, G., Chen, L.C., Murphy, K.P., Yuille, A.L.: Weakly-and semi-supervised learning of a deep convolutional network for semantic image segmentation. In: Proceedings of the IEEE international conference on computer vision. pp. 1742--1750 (2015)

\bibitem{rebuffi2017icarllwfmc}
Rebuffi, S.A., Kolesnikov, A., Sperl, G., Lampert, C.H.: icarl: Incremental classifier and representation learning. In: Proceedings of the IEEE conference on Computer Vision and Pattern Recognition. pp. 2001--2010 (2017)

\bibitem{roy2023rasp}
Roy, S., Volpi, R., Csurka, G., Larlus, D.: Rasp: Relation-aware semantic prior for weakly supervised incremental segmentation. arXiv preprint arXiv:2305.19879  (2023)

\bibitem{shin2017continual}
Shin, H., Lee, J.K., Kim, J., Kim, J.: Continual learning with deep generative replay. Advances in neural information processing systems  \textbf{30} (2017)

\bibitem{sun2020mining}
Sun, G., Wang, W., Dai, J., Van~Gool, L.: Mining cross-image semantics for weakly supervised semantic segmentation. In: Computer Vision--ECCV 2020: 16th European Conference, Glasgow, UK, August 23--28, 2020, Proceedings, Part II 16. pp. 347--365. Springer (2020)

\bibitem{tang2023can}
Tang, L., Xiao, H., Li, B.: Can sam segment anything? when sam meets camouflaged object detection. arXiv preprint arXiv:2304.04709  (2023)

\bibitem{tang2018normalized}
Tang, M., Djelouah, A., Perazzi, F., Boykov, Y., Schroers, C.: Normalized cut loss for weakly-supervised cnn segmentation. In: Proceedings of the IEEE conference on computer vision and pattern recognition. pp. 1818--1827 (2018)

\bibitem{vernaza2017learning}
Vernaza, P., Chandraker, M.: Learning random-walk label propagation for weakly-supervised semantic segmentation. In: Proceedings of the IEEE conference on computer vision and pattern recognition. pp. 7158--7166 (2017)

\bibitem{wang2023scaling}
Wang, D., Zhang, J., Du, B., Tao, D., Zhang, L.: Scaling-up remote sensing segmentation dataset with segment anything model. arXiv preprint arXiv:2305.02034  (2023)

\bibitem{wang2023caption}
Wang, T., Zhang, J., Fei, J., Ge, Y., Zheng, H., Tang, Y., Li, Z., Gao, M., Zhao, S., Shan, Y., et~al.: Caption anything: Interactive image description with diverse multimodal controls. arXiv preprint arXiv:2305.02677  (2023)

\bibitem{wang2020selfseam}
Wang, Y., Zhang, J., Kan, M., Shan, S., Chen, X.: Self-supervised equivariant attention mechanism for weakly supervised semantic segmentation. In: Proceedings of the IEEE/CVF Conference on Computer Vision and Pattern Recognition. pp. 12275--12284 (2020)

\bibitem{wu2018memory}
Wu, C., Herranz, L., Liu, X., Van De~Weijer, J., Raducanu, B., et~al.: Memory replay gans: Learning to generate new categories without forgetting. Advances in Neural Information Processing Systems  \textbf{31} (2018)

\bibitem{wu2019wider}
Wu, Z., Shen, C., Van Den~Hengel, A.: Wider or deeper: Revisiting the resnet model for visual recognition. Pattern Recognition  \textbf{90},  119--133 (2019)

\bibitem{yu2023foundation2023cvpr}
Yu, C., Zhou, Q., Li, J., Yuan, J., Wang, Z., Wang, F.: Foundation model drives weakly incremental learning for semantic segmentation. arXiv preprint arXiv:2302.14250  (2023)

\bibitem{yu2023h2rbox}
Yu, Y., Yang, X., Li, Q., Zhou, Y., Zhang, G., Yan, J., Da, F.: H2rbox-v2: Boosting hbox-supervised oriented object detection via symmetric learning. arXiv preprint arXiv:2304.04403  (2023)

\bibitem{zenke2017continual}
Zenke, F., Poole, B., Ganguli, S.: Continual learning through synaptic intelligence. In: International conference on machine learning. pp. 3987--3995. PMLR (2017)

\bibitem{zhou2016learning}
Zhou, B., Khosla, A., Lapedriza, A., Oliva, A., Torralba, A.: Learning deep features for discriminative localization. In: Proceedings of the IEEE conference on computer vision and pattern recognition. pp. 2921--2929 (2016)

\bibitem{zhu2021prototype}
Zhu, F., Zhang, X.Y., Wang, C., Yin, F., Liu, C.L.: Prototype augmentation and self-supervision for incremental learning. In: Proceedings of the IEEE/CVF Conference on Computer Vision and Pattern Recognition. pp. 5871--5880 (2021)

\end{thebibliography}

\clearpage
\setcounter{page}{1}

\section{Solution to Eq. (\ref{eq: bi-level optimization}) \label{sec:sup solution}}

We focus on the optimization problem 
\begin{equation}
    \begin{split}
        \min_{\mathbf{U},\mathbf{V}}&\quad \mathcal{L}(\theta, \mathbf{U}, \mathbf{V})\\
        {\rm s.t. }& \quad\mathbf{U} \leq \mathbf{1}, \mathbf{V} \leq \mathbf{1}, \mathbf{U} + \mathbf{V} \geq \mathbf{1}.
    \end{split}
\end{equation}
Importantly, three constraints are imposed on $\mathbf{U}$ and $\mathbf{V}$. The first two constraints establish an upper limit on mask fusion, while the third constraint ensures that at least one of the two masks actively contributes to the process. 
This optimization problem can be rewritten as
\begin{align}
        \min_{\mathbf{U},\mathbf{V}}& \sum_{c\in\mathcal{C}^t}\mathcal{L}_{\rm b}\left(\left[\mathbf{U} \odot R(S(\mathbf{x}^t, \mathbf{y}^t), \beta) + \mathbf{V} \odot \mathbf{P}\right]_{c}, \sigma(f_{\theta^t}(\mathbf{x}^t))_{c}\right)\nonumber\\
        {\rm s.t. }& \quad\mathbf{U} \leq \mathbf{1}, \mathbf{V} \leq \mathbf{1}, \mathbf{U} + \mathbf{V} \geq \mathbf{1}.
    \label{eq: uv probelm original}
\end{align}
Note that the logistic function $\sigma(\cdot)$ is an element-wise operator. Therefore, the optimization problem in Eq. (\ref{eq: uv probelm original}) is element-wise independent. To simplify the notation, we represent the elements of each symbol using lowercase letters. Suppose $r = R(S(\mathbf{x}^t, \mathbf{y}^t), \beta)_{ci}$ and $f = \sigma(f_{\theta^t}(\mathbf{x}^t))_{ci}$ for $\forall c\in\mathcal{C}^t, i=(h,w)\in\mathcal{I}$, with $\mathcal{L}_{\rm b}$ being the binary cross-entropy loss, the optimization problem in Eq. (\ref{eq: uv probelm original}) can be transformed into a series of element-wise subproblems as
\begin{equation}
    \begin{split}
        \min_{u,v}& \quad \mathcal{L}(u,v)=\log(\frac{f}{1-f})(u\cdot r + v\cdot p)\\
        {\rm s.t. }& \quad u\leq 1, v\leq 1, u+v \geq 1.
    \end{split}
\end{equation}
Given that the partial derivative of $\mathcal{L}(u,v)$ w.r.t. $u$ or $v$ is a constant, the change of $\mathcal{L}(u,v)$ is monotonic when $u$ or $v$ increases or decreases, which implies that the optimal solution of $u$ and $v$ must be achieved at the boundary. Therefore, the solution to Eq. (\ref{eq: uv probelm original}) can be easily derived as:
\begin{equation}
    \begin{split}
        \mathbf{U}_c &= \phi\left(\mathbb{I}(f_{\theta^t}(\mathbf{x}^t)_c >0 \vee R(S(\mathbf{x}^t, \mathbf{y}^t), \beta)_c \leq \mathbf{P}_c)\right), \\
        \mathbf{V}_c &= \phi\left(\mathbb{I}(f_{\theta^t}(\mathbf{x}^t)_c >0 \vee R(S(\mathbf{x}^t, \mathbf{y}^t), \beta)_c > \mathbf{P}_c)\right).
    \end{split}
    \label{eq: uv solution}
\end{equation}

\section{Algorithm Procedure of Teddy}
We present the complete procedure of Teddy in Alg. \ref{alg}.

\begin{algorithm}[tb]
   \caption{The pseudo code of Teddy}
\begin{algorithmic}
   \STATE {\bfseries Input:} $\mathcal{X}^t$, composed by $\mathbf{x}^t$ with its corresponding image-level annotation $\mathbf{y}^t$ at step $t$, and model $f_{\theta^{t-1}}$ trained at step $t-1$.
   \STATE {\bfseries Output:} The predicted label $y=\{\arg \max_{c\in\mathcal{Y}^t} p_c^i\}^N_{i=1}$, $p_c^i$ is the model prediction of pixel $i$ for class $c$ and $\mathcal{Y}^t$ is the set of seen classes;
   \WHILE{$epoch$ $in$ $num\_epoches$}
   \FOR{($\mathbf{x}^t, \mathbf{y}^t$) in $\mathcal{X}^t$}
   \STATE Compute seed areas $S(\mathbf{x}^t, \mathbf{y}^t)$. 
   \STATE Generate image-level prediction with Global Weighted Pooling and focal penalty based on the seed areas as $\gamma(S(\mathbf{x}^t, \mathbf{y}^t))$.
    \STATE Compute output from the previous model as $f_{\theta^{t-1}}(\mathbf{x}^t)$.
    \STATE Train seed areas with $\mathcal{L}_{cls}$ and $\mathcal{L}_{loc}$ according to Eq. (\ref{eq lcls}, \ref{eq lloc}).
    \IF{$epoch$ $\geq$ 5}
    \STATE Compute model predictions $f_{\theta^t}(\mathbf{x}^t)$.
    \STATE Obtain the binarized predictions for old classes $R(f_{\theta^{t-1}}(\mathbf{x}^t), \alpha)$.
    \STATE Derive $\mathbf{P}$ based on the seed areas.
    \STATE Compute seed areas $S(\mathbf{x}^t, \mathbf{y}^t) = (\mathbf{1}-R(f_{\theta^{t-1}}(\mathbf{x}^t), \alpha))S(\mathbf{x}^t, \mathbf{y}^t)$ based on TME constraint.
    \STATE Compute $\mathbf{U}$ and $\mathbf{V}$ according to Eq. (\ref{eq: uv solution zheng}).
    \STATE Obtain the predictions for new classes $\mathbf{Z}$ according to Eq. (\ref{eq pre z}).
    \STATE Obtain pixel-level supervision $\mathbf{G}$ for all classes according to Eq. (\ref{eq final super}).
    \STATE Train the segmentation model $f_{\theta}$ with $\mathcal{L}_{seg}$ according to Eq. (\ref{eq lseg}).
    \ENDIF
   \ENDFOR
   \ENDWHILE
\end{algorithmic}
\label{alg}
\end{algorithm}

\section{More Experiment Results}
Here we present more complete experiment results including the disjoint setting in Table \ref{tab: disjoint setting results}. Besides, we provide ablation study on 10-5 VOC and COCO-to-VOC settings in Tables \ref{tab:ablation study 10-5}-\ref{tab: mask fusion 10-5}.

\begin{table*}[ht]\footnotesize
 \caption{Results on 15-5 VOC, 10-10 VOC and COCO-to-VOC settings. ``P'' indicates pixel-level labels and ``I'' indicates image-level ones. The best method of utilizing image-level supervision is bold, and the best method using pixel level supervision is underlined. FT is a simple fine-tuning approach, deciding the lower bound of an incremental model, and Joint is trained on all the classed in one step, which can be served as the upper bound.}
    \centering
    \setlength{\tabcolsep}{1.8mm}
    \resizebox{\textwidth}{!}{
    \begin{tabular}{l c | c c c | c c c | c c c | c c c | c c c }
    % \hline\hline
    \toprule
         \multirow{3}{*}{Method} & \multirow{3}{*}{Sup} & \multicolumn{6}{c|}{\textbf{15-5 VOC}} & \multicolumn{6}{c|}{\textbf{10-10 VOC}} & \multicolumn{3}{c}{\textbf{COCO-to-VOC}}\\\cline{3-17}
         & & \multicolumn{3}{c}{Disjoint} & \multicolumn{3}{c|}{Overlap} & \multicolumn{3}{c}{Disjoint} & \multicolumn{3}{c|}{Overlap} & \multicolumn{3}{c}{COCO}  \\ 
         & & 1-15 & 16-20 & All & 1-15 & 16-20 & All & 1-10 & 11-20 & All & 1-10 & 11-20 & All & 1-60 & 61-80 & All \\ 
         
         \hline
         
         FT & P & 8.4 & 33.5 & 14.4 & 12.5 & 36.9 & 18.3 & 7.7 & 60.8 & 33.0 & 7.8 & 58.9 & 32.1 & 1.9 & 41.7 & 12.7 \\ \hline
         
         LWF \cite{li2017learninglwf} & P & 39.7 & 33.3 & 38.2 & 67.0 & 41.8 & 61.0 & 63.1 & 61.1 & 62.2 & \underline{70.7} & 63.4 & 67.2 & 36.7 & \underline{49.0} & \underline{40.3} \\
         
         LWF-MC \cite{rebuffi2017icarllwfmc} & P & 41.5 & 25.4 & 37.6 & 59.8 & 22.6 & 51.0 & 52.4 & 42.5 & 47.7 & 53.9 & 43.0 & 48.7  & - & - & - \\

         ILT \cite{michieli2019incrementalilt} & P & 31.5 & 25.1 & 30.0 & 69.0 & 46.4 & 63.6 & \underline{67.7} & \underline{61.3} & \underline{64.7} & 70.3 & 61.9 & 66.3 & \underline{37.0} & 43.9 & 39.3 \\

         CIL \cite{klingner2020classcil} & P & 42.6 & 35.0 & 40.8 & 14.9 & 37.3 & 20.2 & 37.4 & 60.6 & 48.8 & 38.4 & 60.0 & 48.7 & - & - & - \\
 
         MiB \cite{cermelli2020modelingmib} & P & 71.8 & 43.3 & 64.7 & 75.5 & 49.4 & 69.0 & 66.9 & 57.5 & 62.4 & 70.4 & 63.7 & 67.2 & 34.9 & 47.8 & 38.7  \\

         PLOP \cite{douillard2021plop} & P & 71.0 & 42.8 & 64.3 & \underline{75.7} & 51.7 & \underline{70.1} & 63.7 & 60.2 & 63.4 & 69.6 & 62.2 & 67.1 & 35.1 & 39.4 & 36.8 \\

         SDR \cite{michieli2021continualsdr} & P & \underline{73.5} & 47.3 & \underline{67.2} & 75.4 & 52.6 & 69.9 & 67.5 & 57.9 & 62.9 & 70.5 & \underline{63.9} & \underline{67.4} & - & - & - \\

         RECALL \cite{maracani2021recall} & P & 69.2 & \underline{52.9} & 66.3 & 67.7 & \underline{54.3} & 65.6 & 64.1 & 56.9 & 61.9 & 66.0 & 58.8 & 63.7 & - & - & - \\

        \hline
        
         CAM \cite{zhou2016learning} & I & 69.3 & 26.1 & 59.4 & 69.9 & 25.6 & 59.7 & 65.3 & 41.3 & 54.5 & 70.8 & 44.2 & 58.5 & 30.7 & 20.3 & 28.1 \\

         SEAM \cite{wang2020selfseam} & I & 71.0 & 33.1 & 62.7 & 68.3 & 31.8 & 60.4 & 65.1 & 53.5 & 60.6 & 67.5 & 55.4 & 62.7 & 31.2 & 28.2 & 30.5 \\

         SS \cite{araslanov2020singless} & I & 71.6 & 26.0 & 61.5 & 72.2 & 27.5 & 62.1 & 60.7 & 25.7 & 45.0 & 69.6 & 32.8 & 52.5 & 35.1 & 36.9 & 35.5   \\

         EPS \cite{lee2021railroadeps} & I & 72.4 & 28.5 & 65.2 & 69.4 & 34.5 & 62.1 & 64.2 & 54.1 & 60.6 & 69.0 & 57.0 & 64.3 & 34.9 & 38.4 & 35.8\\
         
         WILSON \cite{cermelli2022incrementalwilson} & I & 73.6 & 43.8 & 67.3 & 74.2 & 41.7 & 67.2 & 64.5 & 54.3 & 60.8 & 70.4 & 57.1 & 65.0 & 39.8 & 41.0 & 40.6\\
         
         \textbf{Teddy} & I & \textbf{74.5} & \textbf{48.1} & \textbf{69.0} & \textbf{77.6} & \textbf{51.4} & \textbf{72.0} & \textbf{65.4} & \textbf{55.2} &  \textbf{61.7} & \textbf{71.2} & \textbf{59.4} & \textbf{66.5} & 
         \textbf{40.6} & \textbf{41.8} & \textbf{41.5} 
         \\\hline
        
        Joint & P & 75.5 & 73.5 & 75.4 & 75.5 & 73.5 & 75.4 & 76.6 & 74.0 & 75.4 & 76.6 & 74.0 & 75.4 & - & - & -\\
         \bottomrule
    \end{tabular}
    }

    \label{tab: disjoint setting results}
\end{table*}

\begin{table}[ht]\footnotesize
    \centering
    \setlength{\tabcolsep}{1.3mm}
    \caption{Ablation study. ``OB'' stands for prediction binarization for old classes, ``TME'' stands for tendency-driven mutual exclusivity and ``PF'' stands for prediction fusion for new classes.}
    \begin{tabular}{c c c c|c c c | c c c }
    \toprule
    \multirow{2}{*}{Row} & \multicolumn{2}{c}{TME} & \multirow{2}{*}{PF} & \multicolumn{3}{c|}{\textbf{10-5 VOC}} & \multicolumn{3}{c}{\textbf{COCO-to-VOC}} \\
    & OB & w/o OB & & 1-10 & 11-20 & All & 1-60 & 61-80 & All \\\hline
    
    1 & & & & 66.8 & 46.5 & 58.1 & 39.8 & 41.0 & 40.6 \\
    
    2 & & & \checkmark & 67.0 & 50.7 & 60.3 & 40.6 & 39.7  & 40.9\\

    3 & & \checkmark & & 67.2 & 51.6 & 60.8 &  40.9 & 39.8 & 41.1 \\

    4 & \checkmark & & & 67.6 & 52.0 & 61.2 & 41.1 & 39.8 & 41.3  \\

    5 & & \checkmark & \checkmark &  67.2 & 51.8 & 60.9  & 41.1 & 40.0 & 41.3 \\

    \hline
    
    6 & \checkmark &  & \checkmark & 68.9 & 51.7 & 61.7 & 40.6 & 41.8 & 41.5\\
    
    \bottomrule
    \end{tabular}
    \label{tab:ablation study 10-5}
\end{table}

\begin{table*}[ht]
    \centering
    \setlength{\tabcolsep}{1.3mm}
    \caption{Effectiveness of prediction fusion for new classes through optimization.}
    
    \begin{tabular}{c c|c c c | c c c }
    \toprule
    \multirow{2}{*}{$\mathbf{U}$} & \multirow{2}{*}{$\mathbf{V}$} & \multicolumn{3}{c|}{\textbf{10-5 VOC}} & \multicolumn{3}{c}{\textbf{COCO-to-VOC}} \\
     & & 1-10 & 11-20 & All & 1-60 & 61-80 & All\\\hline
     
     0.25 & 0.75 & 66.6 & 46.3 & 58.0 & 40.6 & 37.0 & 40.2\\
     
     0.50 & 0.50 & 67.1 & 51.2 & 60.5 & 40.6 & 39.9 & 40.9 \\
     
     0.75 & 0.25 & 67.0 & 50.8 & 60.3 & 40.4 & 38.1 & 40.5 \\\hline
    \multicolumn{2}{c|}{Optimization}& 68.9 & 51.7 & 61.7 & 40.6 & 41.8 & 41.5\\
    \bottomrule
    \end{tabular}
    \label{tab: mask fusion 10-5}
\end{table*}

\begin{table}[ht]
    \centering
     \caption{Sensitivity analysis of Teddy on $\alpha$ and $\beta$ under 15-5 VOC setting.}
    \setlength{\tabcolsep}{1.8mm}
    \begin{tabular}{c c | c c c | c c c}
     \toprule
         \multirow{2}{*}{$\alpha$} & \multirow{2}{*}{$\beta$} & \multicolumn{3}{c|}{Disjoint} & \multicolumn{3}{c}{Overlap} \\
         & & 1-15 & 16-20 & All & 1-15 & 16-20 & All \\\hline

         0.9 & \multirow{5}{*}{0.5} & 74.3 & 46.0 & 68.4 & 77.2 & 51.1 & 71.6\\
         0.8 & & 74.5 & 48.1 & 69.0 & 77.1 & 51.3 & 71.6 \\
         0.7 & & 74.5 & 48.1 & 69.0 & 77.1 & 51.1 & 71.5 \\
         0.6 & & 74.3 & 48.1 & 68.8 & 77.1 & 51.2 & 71.5\\
         0.5 & & 74.1 & 47.9 & 68.6 & 77.3 & 49.3 & 71.3 \\\hline

         \multirow{5}{*}{0.8} & 0.9 & 74.3 & 45.4 & 68.3 & 77.6 & 50.2 & 71.7 \\
          & 0.8 & 74.4 & 45.9 & 68.4 & 77.0 & 50.7 & 71.4 \\
          & 0.7 & 74.4 & 46.0 & 68.5 & 76.9 & 50.5 & 71.3 \\
          & 0.6 & 74.4 & 45.9 & 68.4 & 76.8 & 50.2 & 71.2 \\
          & 0.5 & 74.5 & 48.1 & 69.0 & 77.6 & 51.4 & 72.0 \\

    \bottomrule
    \end{tabular}
    \label{tab: sensi}
\end{table}

\section{Sensitivity Analysis}
We have conducted a sensitivity analysis for the hyper-parameters $\alpha$ and $\beta$, which govern the ratio used during the binarization of predictions based on SAM. The results of this analysis are presented in Tab. \ref{tab: sensi}. Notably, the results exhibit a high degree of stability even when $\alpha$ varies over a wide range. This stability serves as validation for the accuracy and consistency of predictions for the old classes generated by the previous model.

\iffalse
\section{Other Details}
Following \cite{cermelli2022incrementalwilson}, we also adopt the self-supervised segmentation loss on the localizer. For more details on this, please refer to \cite{cermelli2022incrementalwilson}. 

We opted not to compare Teddy with another WILSS method, FWILSS \cite{yu2023foundation2023cvpr}. Our rationale lies in our belief that there are fundamental flaws in \cite{yu2023foundation2023cvpr}. Specifically, the authors retain images from the $t-1$ step along with their corresponding ground-truth pixel-level supervisions. In the $t$ step, they directly use these images and labels as a means of augmentation to train the incremental model. Even within the context of ILSS, such an approach can be considered a severe case of data leakage, which is in direct violation of the foundational principles of incremental learning.
\fi

\section{Discussion on Semantic Foundation Model}
Recently, many large models have been proposed in semantic segmentation. They achieve impressive performances, especially for their adeptness in weakly supervised and even zero-shot learning scenarios. Although it may seem that they can do anything in the computer vision community, this does not devalue research in specific segmentation challenges. These tasks often require fine-tuning of semantic foundation models for unique goals - a process demanding significant resources. Additionally, exploring specific scenarios can further advance these models, driving the evolution of research in the computer vision community.

Moreover, while models like SemanticSAM \cite{li2023semanticsam} are proficient in open-set segmentation, they assume all classes are known during one step training. The need for these models to adapt to unknown labels underscores the importance of incremental learning, enabling them to integrate new information naturally, akin to human learning. Given the high costs and efforts needed for pixel-level annotations, WILSS is highly motivated and relevant to the community.

\end{document}